\documentclass[10pt,twocolumn,letterpaper]{article}

\usepackage{times}
\usepackage{epsfig}
\usepackage{graphicx}
\usepackage{amsmath}
\usepackage{amssymb}

\usepackage{booktabs}
\usepackage{makecell}
\usepackage{tikz}
\usepackage[english]{babel}
\usepackage{amsthm}
\usepackage{pifont}
\newcommand{\cmark}{\ding{51}}
\newcommand{\xmark}{\ding{55}}

\newtheorem{theorem}{Theorem}[section]
\newtheorem{lemma}[theorem]{Lemma}

\definecolor{forestgreen}{RGB}{34,139,34}
\usepackage{colortbl}
\usepackage{multirow}

\newcommand{\vB}{\mathbf{B}}
\newcommand{\vI}{\mathbf{I}}
\newcommand{\vM}{\mathbf{M}}
\newcommand{\vF}{\mathbf{F}}
\newcommand{\vQ}{\mathbf{Q}}
\newcommand{\vK}{\mathbf{K}}
\newcommand{\vV}{\mathbf{V}}
\newcommand{\vX}{\mathbf{X}}
\newcommand{\vW}{\mathbf{W}}
\newcommand{\vO}{\mathbf{O}}
\newcommand{\vR}{\mathbb{R}}

\newcommand{\vc}{\mathsf{c}}
\newcommand{\vh}{\mathsf{h}}

\newcommand{\vw}{\mathsf{w}}
\newcommand{\vbeta}{\mathsf{\beta}}
\newcommand{\vp}{\mathsf{p}}
\newcommand{\vx}{\mathsf{x}}
\newcommand{\vq}{\mathsf{q}}
\newcommand{\vf}{\mathbf{f}}
\newcommand{\vt}{\mathbf{t}}

\newcommand{\vL}{\mathcal{L}}

\usepackage{cvpr}              

\definecolor{cvprblue}{rgb}{0.21,0.49,0.74}
\usepackage[pagebackref,breaklinks,colorlinks,citecolor=cvprblue]{hyperref}

\def\paperID{1706} 
\def\confName{CVPR}
\def\confYear{2024}

\title{LDP: Language-driven Dual-Pixel Image Defocus Deblurring Network}

\author{Hao Yang$^{1}$, \quad Liyuan Pan$^{1}$, \quad Yan Yang$^{2}$, \quad Richard Hartley$^{2}$, \quad and \quad Miaomiao Liu$^{2}$ \\
$^{1}$Beijing Institute of Technology \quad $^{2}$ The Australian National University \\
{\tt\small \{hao.yang, liyuan.pan\}@bit.edu.cn, \quad \{yan.yang, richard.hartley, miaomiao.liu\}@anu.edu.au}
}

\begin{document}
\maketitle
\begin{abstract}
Recovering sharp images from dual-pixel (DP) pairs with disparity-dependent
blur is a challenging task.~Existing blur map-based deblurring methods 
have demonstrated promising results. In this paper, we propose, to the best of our knowledge, the first framework that introduces the contrastive language-image pre-training framework (CLIP) to accurately estimate the blur map from a DP pair unsupervisedly. To achieve this, we first carefully design text prompts to enable CLIP to understand blur-related geometric prior knowledge from the DP pair. Then, we propose a format to input a stereo DP pair to CLIP without any fine-tuning, despite the fact that CLIP is pre-trained on monocular images. Given the estimated blur map, we introduce a blur-prior attention block, a blur-weighting loss, and a blur-aware loss to recover the all-in-focus image. Our method achieves state-of-the-art performance in extensive experiments (see Fig.~\ref{fig:teaser}).  
\end{abstract}

\section{Introduction}

Defocus deblurring from a single image is an ill-posed problem \cite{ICCV21_KPAC,quan2023neumann,li2023learning,quan2023single}. Compared to methods that restore an all-in-focus image directly \cite{abuolaim2020defocus,abuolaim2021learning}, blur map-based deblurring methods have demonstrated promising results \cite{lee2021iterative,pan2021dual}. To simplify the blur removal process, several works \cite{xin2021defocus,liang2021bambnet,yang2023k3dn} have utilized recent innovative dual-pixel (DP) sensors \cite{choi2023exploring} to estimate blur maps, since the DP sensor captures a two-view image pair (a DP pair) in a single shot. The DP pair can be treated as images from a two-sample light field camera or a stereo system with a tiny baseline \cite{xin2021defocus,punnappurath2020modeling}, which is beneficial for disparity estimation \cite{zhang2020du2net,abuolaim2020defocus,kim2023spatio,monin2023continuous}. Based on the DP image pair formation process, the disparity is geometrically related to the blur map, \ie, the degree of defocus blur proportional to disparity magnitudes~\cite{garg2019learning}. 
 
For estimating the blur map (defocus map or disparity map), previous works either require extra data \cite{pan2021dual, abuolaim2022improving}, such as synthetic data, for supervision signals, or pre-calibrated blur kernels \cite{xin2021defocus}. To avoid using extra information, several works explored unsupervised blur map estimation frameworks~\cite{lee2021iterative, liang2021bambnet}. Lee \etal \cite{lee2021iterative} implicitly estimate the disparity of the left/right views using a spatial transformer, yet only in feature space, as well as Yang \etal \cite{yang2023k3dn}. Liang \etal \cite{liang2021bambnet} use fixed network branches for layer-wise defocus map estimation and deblurring, where the limited network branch number restricts the potential of its accuracy. 

Recently, the contrastive language-image pre-training framework (CLIP) 
\cite{radford2021learning} has accomplished extraordinary success for vision tasks, such as semantic segmentation, object detection, and 3D point cloud understanding~\cite{rao2022denseclip,shi2023edadet,zhang2022pointclip}. Therefore, a question naturally arises,
~\emph{can we avoid the cost of collecting data and model designing, to estimate the blur 
map unsupervisedly by using {\rm CLIP}?} Unfortunately, using semantic knowledge from CLIP to handle 
low-level vision tasks is underexplored.

\begin{figure}[!t]
    \centering
    \includegraphics[width=\linewidth]{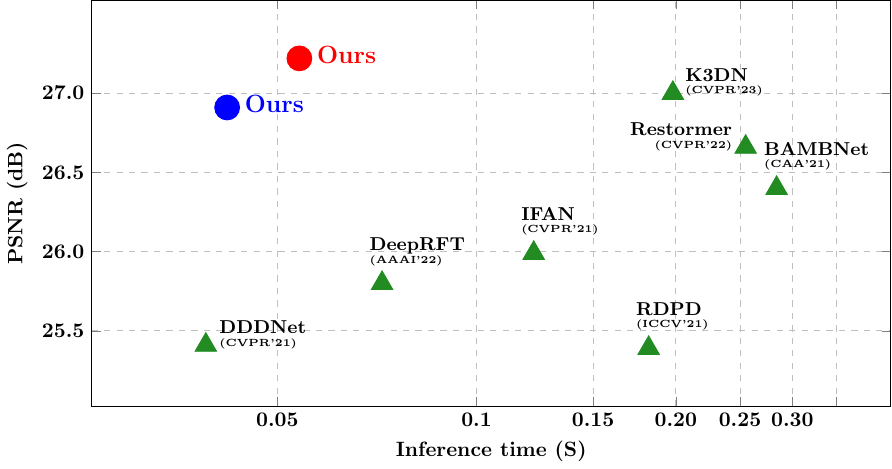}
    \vspace{-6mm}
    \caption{\small \it 
Comparison of performance in terms of PSNR and Inference time. Our small and large models are respectively denoted by the \textcolor{blue}{Blue} and \textcolor{red}{Red} cycles. The \textcolor{forestgreen}{Green} triangles represent previous state-of-the-art methods, and each of them is annotated with the method name, publication venue, and year. This figure is best viewed in color on the screen.
    }
    \vspace{-5mm}
    \label{fig:teaser}
\end{figure}

In this paper, for the first time, we explore the relationship between language and blur map estimation by applying CLIP.~Note that CLIP is trained to match a monocular image with its corresponding language descriptions. 
Given a blurred image, it is intuitive to match it with a corresponding text \texttt{[A blurry image.]}, which allows the language model to find the blurred regions in an image. However, it is hard to quantize the blur level of an image without knowing its associated latent sharp one.

Fortunately, we can leverage the capabilities of DP sensors.~With DP pairs, blur quantization becomes possible without the need for its associated sharp image. 
The formation of DP ensures no disparity between left/right views for pixels in sharp regions, while a noticeable disparity exists for pixels in blurred regions. 
In essence, the detectable differences (disparities) in left/right views of a DP pair directly indicate the degree of defocus blurs. Consequently, we can quantify the variation between the left and right views, treating it as a valuable cue for estimating a blur map.  

However, CLIP still finds it challenging to simultaneously explore stereo DP pairs with left/right view images. 
To address this issue, we propose to explore the DP pair and design an `image and text' format to better capture the blurs in the image.
Specifically, we propose to create a new image from the DP pair and~\emph{translate} the description about the `blurriness' of the image to measure the `symmetry' of the newly formed image{, which is based on the DP image formation process}. The freshly designed image-text format allows the CLIP to estimate the blur map easily (see \cref{sec:det}).

Given the estimated blur map, we then restore the all-in-focus image from the DP pair. By observing that the attention map of the self-attention mechanism \cite{dosovitskiy2020image} can be used in the dynamic DP deblurring process, we introduce a blur prior attention (BPA) that modulates the attention map with the estimated blur map which provides prior knowledge of deblurring kernels (see \cref{sec:net}).    

To regulate the restoration, we specifically design two losses that are based on CLIP.
Considering the network should focus more on heavily blurred regions, we propose a blur-weighting loss that imposes an adaptive penalty proportional to the blur degrees provided by the estimated blur map from CLIP.
Then, to guarantee the restored image is sharp, we use CLIP to re-check the restored image and formulate a blur-aware loss (see \cref{sec:los}).  

Our contributions are summarised as follows:

\vspace{-.3em}
\begin{itemize}
    \setlength\itemsep{-.2em}
    \item We propose a {\bf L}anguage-driven {\bf DP} ({\bf LDP}) defocus deblurring framework, by exploring the potential of CLIP in the low-level vision task.
    \item We design an image-text format for blur map estimation based on the geometric relationship between the blur and disparity of the DP pair.
    \item We propose a blur prior attention (BPA) block, a blur-weighting loss, and a blur-aware loss to encourage sharp restorations of the DP pair.
\end{itemize}
\vspace{-.3em}

We evaluate the proposed framework extensively on standard benchmark datasets. Though using the large-scale foundation model, CLIP, for blur map estimation, our method achieves {Pareto-optimality} between deblurring performance and inference speed (refer to Fig.~\ref{fig:teaser} for an overview of the comparison). Our codes and models will be released to facilitate reproducible research.

\begin{figure*}[!t]
    \centering
    \includegraphics[width=0.85\textwidth]{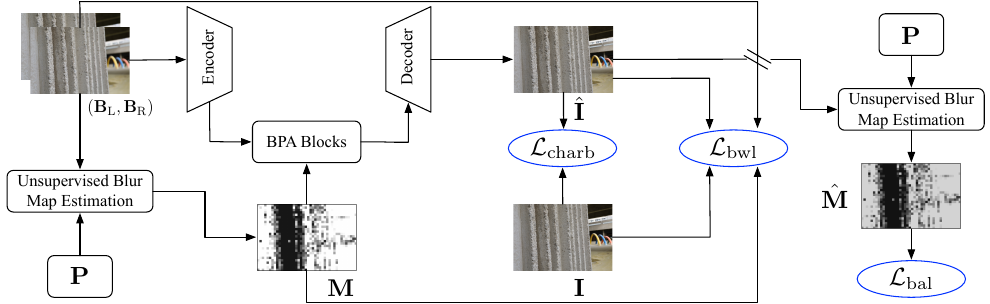}
    \vspace{-2mm}
    \caption{\small \it Overall architecture. Given a DP pair $(\vB_{\rm L}, \vB_{\rm R})$, we first generate a blur map $\vM$ by using our unsupervised blur map estimation strategy which is based on the CLIP, and the nsembling format is used. We then forward the DP pair $(\vB_{\rm L}, \vB_{\rm R})$ to our deblurring backbone composed of an encoder and a decoder. Meanwhile, we have multiple BPA blocks, using the blur map $\vM$ as deblurring kernel prior to refine the intermediate backbone feature embeddings. In the following, we have the restoration $\hat{\vI}$. The deblurring network is optimized with three losses, charbonnier loss $\vL_{\text{charb}}$, blur weighting loss $\vL_{\text{bwl}}$, and blur aware loss $\vL_{\text{bal}}$. $\vL_{\text{charb}}$ penalizes deviations from $\vI$ in the resortation $\hat{\vI}$. $\vL_{\text{bwl}}$ and $\vL_{\text{bal}}$ regularize the restoration by 
    imposing an adaptive penalty on the blurred regions of the DP pair, and leveraging unsupervised blur map estimation strategy to encourage the restoration to be sharp, \ie, the blur map $\hat{\vM}$ for the restoration $\hat{\vI}$ is zero value filled.    
    }
    \label{fig:arch}
    \vspace{-5mm}
\end{figure*}

\section{Related Work}

\paragraph{DP Defocus Deblurring.} DP-based deblurring methods can be divided into single- and multi-task manners. For single-task based deblurring \cite{abuolaim2020defocus,abuolaim2021learning,abuolaim2022improving}, methods apply different kernels at different regions of the DP pair to handle the spatially varying defocus deblur. This can usually be achieved by using parallel atrous convolutions \cite{ICCV21_KPAC},  attention mechanisms \cite{zamir2022restormer}, or deformable convolutions \cite{li2022learning}. 
For multi-task deblurring methods, the blur/defocus/disparity map is auxiliarily estimated to benefit DP defocus deblurring. Pan \etal \cite{pan2021dual} and Xin \etal \cite{xin2021defocus} use the estimated DP disparity to regularise the restoration through reblurring; however, with extra data or pre-calibrate kernel. Several works explored unsupervised defocus map (or disparity) estimation. Lee \etal \cite{lee2021iterative} and Yang \etal \cite{yang2023k3dn} estimate defocus disparity only in feature space without explicitly showing a defocus map. Liang \etal \cite{liang2021bambnet} optimize a multi-channel defocus map based on \cite{punnappurath2020modeling}. 
However, hyperparameters, such as the defocus threshold and the number of channels, are preset, limiting blur map estimation accuracy. 
Unlike past works, by transferring the blur-related geometric prior knowledge into semantic prompts, we unsupervisedly estimate the blur map from CLIP to benefit defocus deblurring.

\vspace{-5mm}
\paragraph{Language-driven Vision.} Common objectives in visual-language learning are aligning the text with a paired image in the embedding space \cite{radford2021learning,liang2023iterative}. The CLIP, one of the milestone achievement in the area, demonstrate that language-independent computer vision tasks can be substantially benefited from visual-language learning, by achieving SOTA zero-shot classification performance in the ImageNet-1K benchmark \cite{deng2009imagenet}. Successive works have extended the zero-shot learning capability of CLIP to semantic segmentation \cite{rao2022denseclip,zhou2023zegclip}, monocular depth estimation \cite{zhang2022can,auty2023learning,hu2023learning}, and point clouds \cite{zhang2022pointclip,huang2023clip2point}. 
However, none of the existing works explores the zero-shot learning capability of CLIP in stereo/DP pairs. 
This paper proposes the first approach to use the monocular image-trained CLIP in zero-shot blur map segmentation of DP pairs.

\section{Method}

Let us now introduce our language-driven dual-pixel image defocus deblurring framework. Given a defocus blurred DP pair $(\vB_{\rm L}, \vB_{\rm R})$, we aim to restore an all-in-focus image $\vI$, where $\vB_{\rm L}$ and $\vB_{\rm R}$ are the left and right view of a DP pair and $\vB_{\rm L},\vB_{\rm R},\vI \in \vR^{\vh \times \vw \times 3}$. Here, $\vh$ and $\vw$ define the image height and width, respectively. 

We first design a DP-aware prompt to adapt CLIP to output a blur map, $\vM \in \vR^{\vh \times \vw \times 1}$, which quantizes disparity unsupervisedly between $\vB_{\rm L}$ and $\vB_{\rm R}$. Then, we propose a BPA block based network which employs the blur map $\vM$ as deblurring priors to restore 
a sharp image $\hat{\vI}$ from 
$(\vB_{\rm L}, \vB_{\rm R})$.
To regularize the restoration of $\hat{\vI}$, a blur-weighting loss $\vL_{\text{bwl}}$ and a blur-aware loss $\vL_{\text{bal}}$ are also introduced that were underpinned by our unsupervised blur map estimation strategy. The overall framework is shown in Fig.~\ref{fig:arch}. Details are presented below.

\subsection{Unsupervised Blur Map Estimation}
\label{sec:det}
\paragraph{Preliminary.} CLIP respectively embeds an image and a text description into a common semantic space of dimension $\vc$ with an image encoder and a text encoder, obtaining image embedding $\vf \in \vR^{\vc} $ and text embedding $\vt \in \vR^{\vc}$. By optimizing a contrastive loss, the embeddings of the aligned image and text pair, $\Omega_a$, stay similar, \ie, maximizing the semantic similarity by $\max_{\vf, \vt\in \Omega_a} \tt{sim}(\vf, \vt)$, where $\tt{sim}(\cdot,\cdot)$ is the cosine similarity metric. Meanwhile, the semantics of misaligned pairs, $\Omega_{na}$ are pushed away from each other, \ie, $\min_{\vf, \vt\in\Omega_{na}} \tt{sim}(\vf, \vt)$. With comprehensive training data, CLIP can be leveraged to align images with pre-defined prompt templates composed of open-world text descriptions for zero-shot learning \cite{radford2021learning}.

For a DP pair, the disparity $d$ for a pixel in the left view is defined
by its location $(i,j)$ and that of its corresponding pixel at $(i,k)$
in the right view. 

In the literature, this DP disparity \cite{punnappurath2020modeling} is defined as
\begin{align}
   d=j-k= A\bigl(r(z)\bigr) \cdot B(z)\ , ~ \vB_{\rm L} (i,j) = \vB_{\rm R}(i,k) \ , \label{eq:disparity}
\end{align}
where $\vB_{\rm L} (i,j)$ and $\vB_{\rm R}(i,k)$ are the intensity of the two matching points, {$r(\cdot)$ is a function of the radius of the circle of confusion}, $z$ is the point depth, and $A(r(z))$ and $B(z)$ are lens-dependent constants \cite{pan2021dual,xin2021defocus,abuolaim2021learning,garg2019learning}. 
\vspace{-5mm}
\paragraph{Analysis.} Leveraging the CLIP, zero-shot blur map estimation can be generally divided into three steps. {\it i) Image Encoding.} Considering the image encoder of the CLIP is trained with monocular RGB images, a function $\eta(\cdot, \cdot)$ is required to create a new monocular image
from the DP image pair $(\vB_{\rm L}, \vB_{\rm R})$. To obtain the dense embedding for the blur map estimation task, we follow \cite{zhang2022can} to pop out the final pooling layer from the image encoder, yielding an embedding matrix $\vF \in \vR^{\vh_{\mathsf{s}} \times \vw_{\mathsf{s}} \times \vc}$, where $\vh_{\mathsf{s}}$ and $\vw_{\mathsf{s}}$ are the height and width for embedding of the newly created image, respectively.
Mathematically, we have 
\begin{align}
    \vF = \text{ImageEncoder}\bigl(\eta(\vB_{\rm L},\vB_{\rm R})\bigr) \ .
\end{align}
Each pixel of $\vF$ has encoded the local semantic information for a corresponding region in $\eta(\vB_{\rm L},\vB_{\rm R})$. 
{\it ii) Text Encoding.} A text encoder is used to embed the semantics of a prompt $\vp$ describing the blur into the text embedding,
\begin{align}
    \vt = \text{TextEncoder}(\vp) \ . 
\end{align} 
{\it iii) Blur map estimation.} Let $\vF(i,j)$ be a $\vc$-dimensional embedding at position $(i,j)$ of $\vF$, $i \in \{1,\cdots,\vh_\mathsf{s}\}$, and $j \in \{1,\cdots,\vw_{\mathsf{s}}\}$. The embedding $\vF(i,j)$ and the blur semantic-based text embedding $\vt$ are expected to be similar if the pixel at $(i,j)$ is blurred. Therefore, the blur mask $\vM_{\vF}$ of $\eta(\vB_{\rm L},\vB_{\rm R})$ 
is obtained by 
\begin{align}
    \vM_{\vF}(i,j) = \sigma\bigl(\texttt{sim}(\vF(i,j),\vt)\bigr) \ . \label{eq:maskbar}
\end{align}
Note, $\sigma(\cdot)$ normalizes the cosine similarity $\tt{sim}(\cdot,\cdot)$ output to a probability space, to reflect the degree of blur at the position. Here, $\vM_{\vF}$ is sampled to have the same resolution as $\vB_{\rm L}$ or $\vB_{\rm R}$, returning $\vM \in \vR^{\vh \times \vw \times 1}$.

Given the analysis above, our goal is to develop text prompts $\vp$, while adapting the function $\eta(\cdot,\cdot)$, to enable the use of semantic information from CLIP to generate the blur map
$\vM$ in an unsupervised manner.
{In particular, we explore the blur-aware, DP-aware, and ensembling formats with detailed presented below.}

\begin{figure}[!t]
    \centering
    \begin{minipage}{0.24\linewidth}
    \begin{minipage}{\linewidth}
       \centering
       \includegraphics[width=\linewidth]{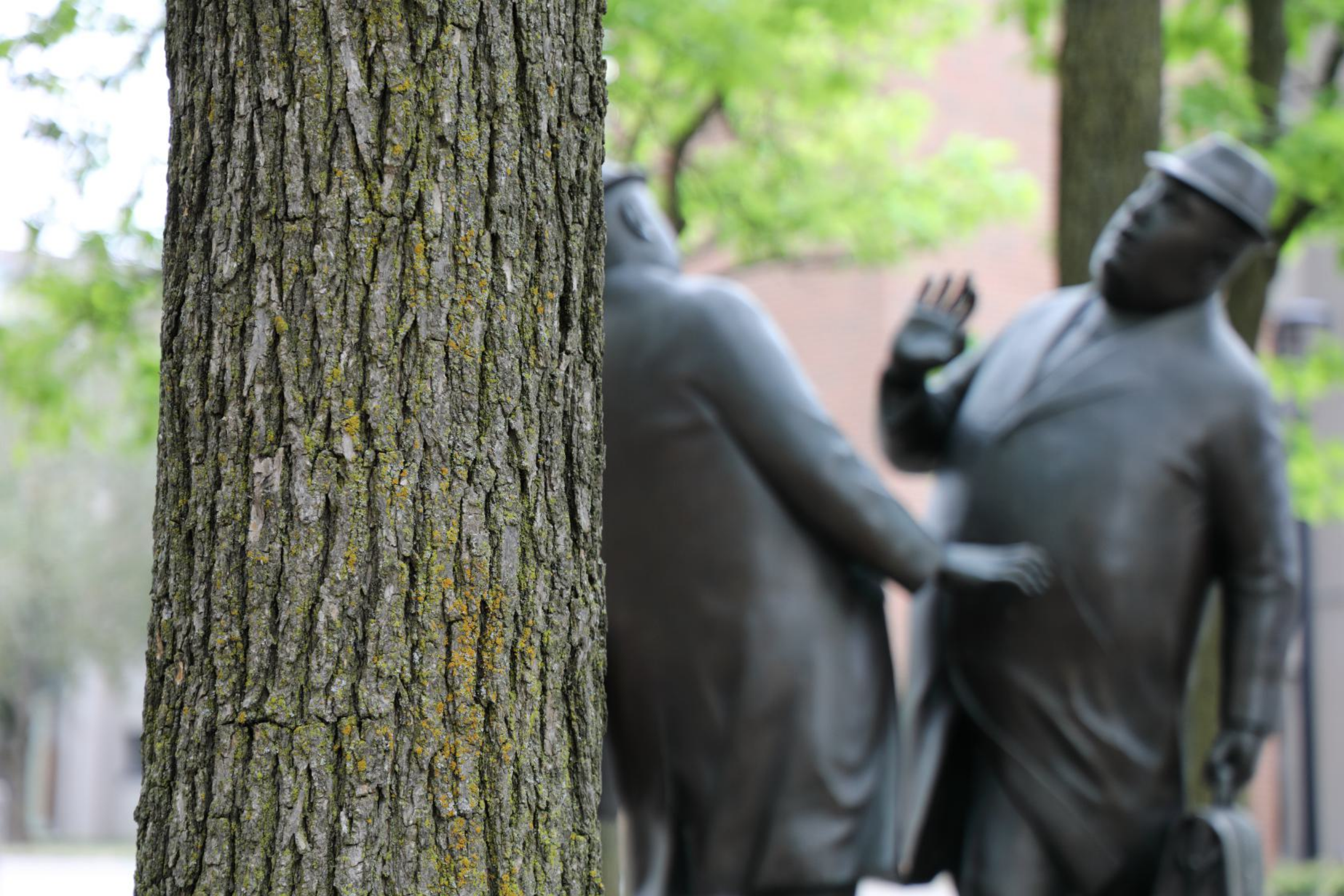}
    \end{minipage}
    \begin{minipage}{\linewidth}
       \centering
       \includegraphics[width=\linewidth]{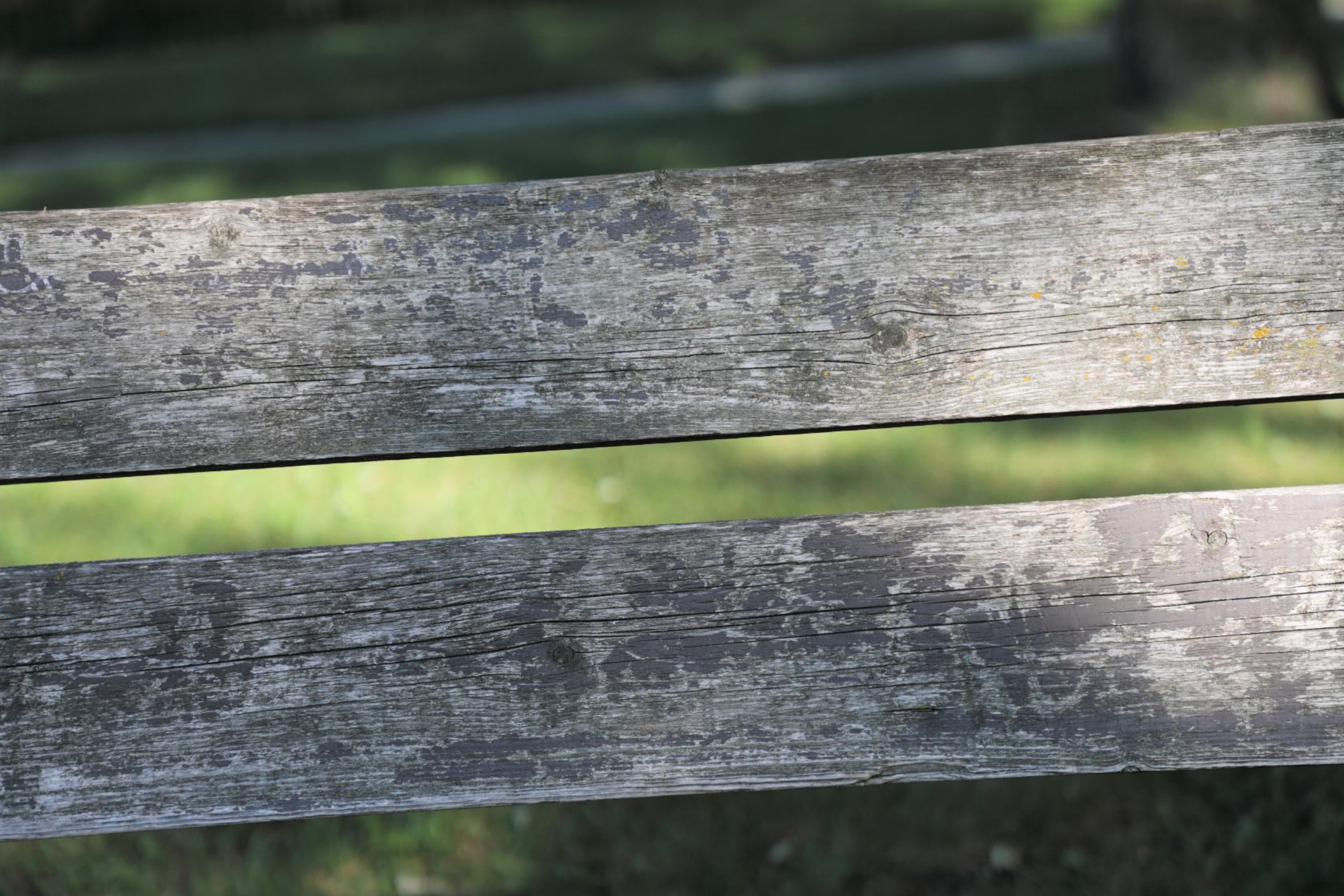}
       \subcaption{}
    \end{minipage}
    \end{minipage}
        \begin{minipage}{0.24\linewidth}
    \begin{minipage}{\linewidth}
       \centering
       \includegraphics[width=\linewidth]{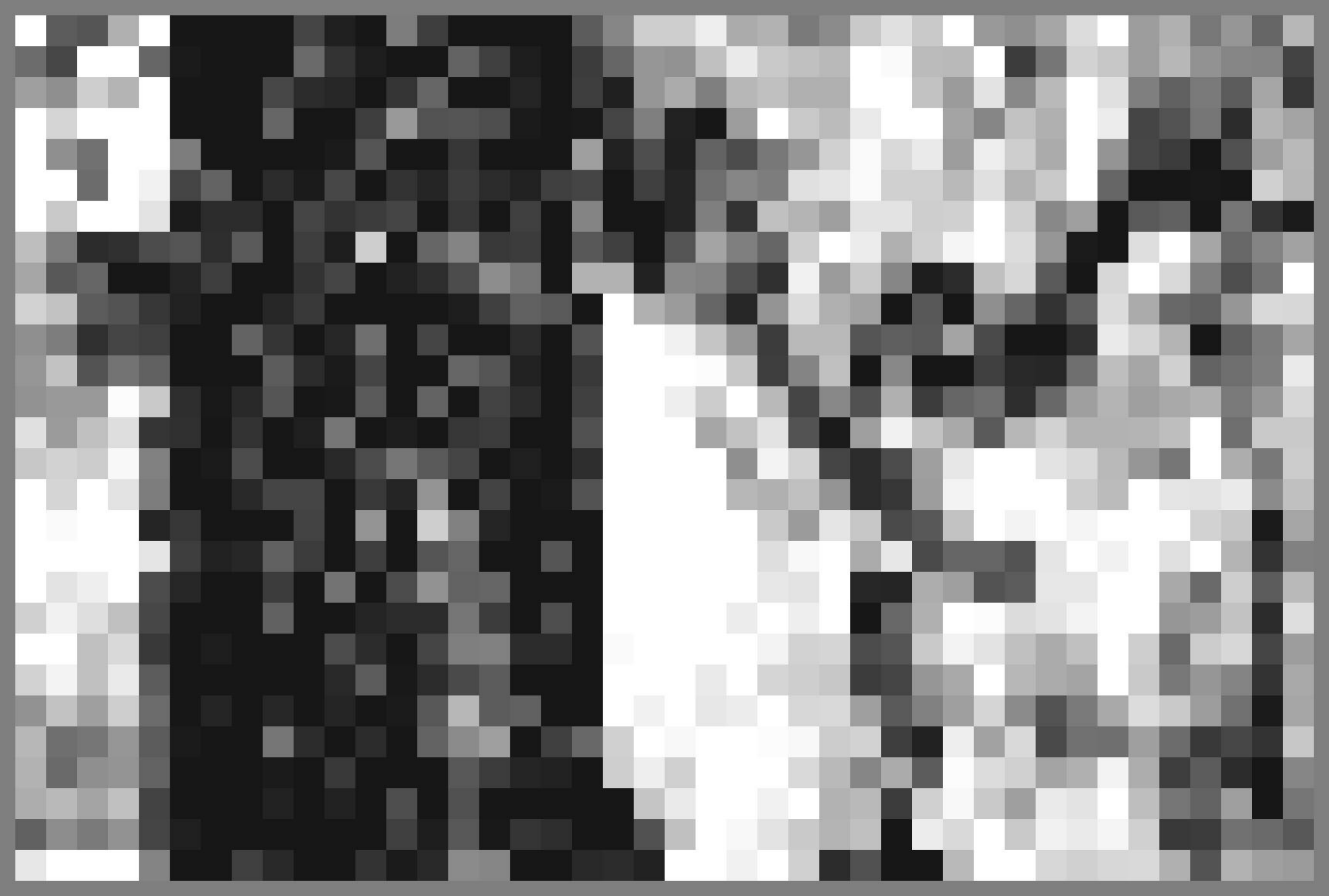}
    \end{minipage}
    \begin{minipage}{\linewidth}
       \centering
       \includegraphics[width=\linewidth]{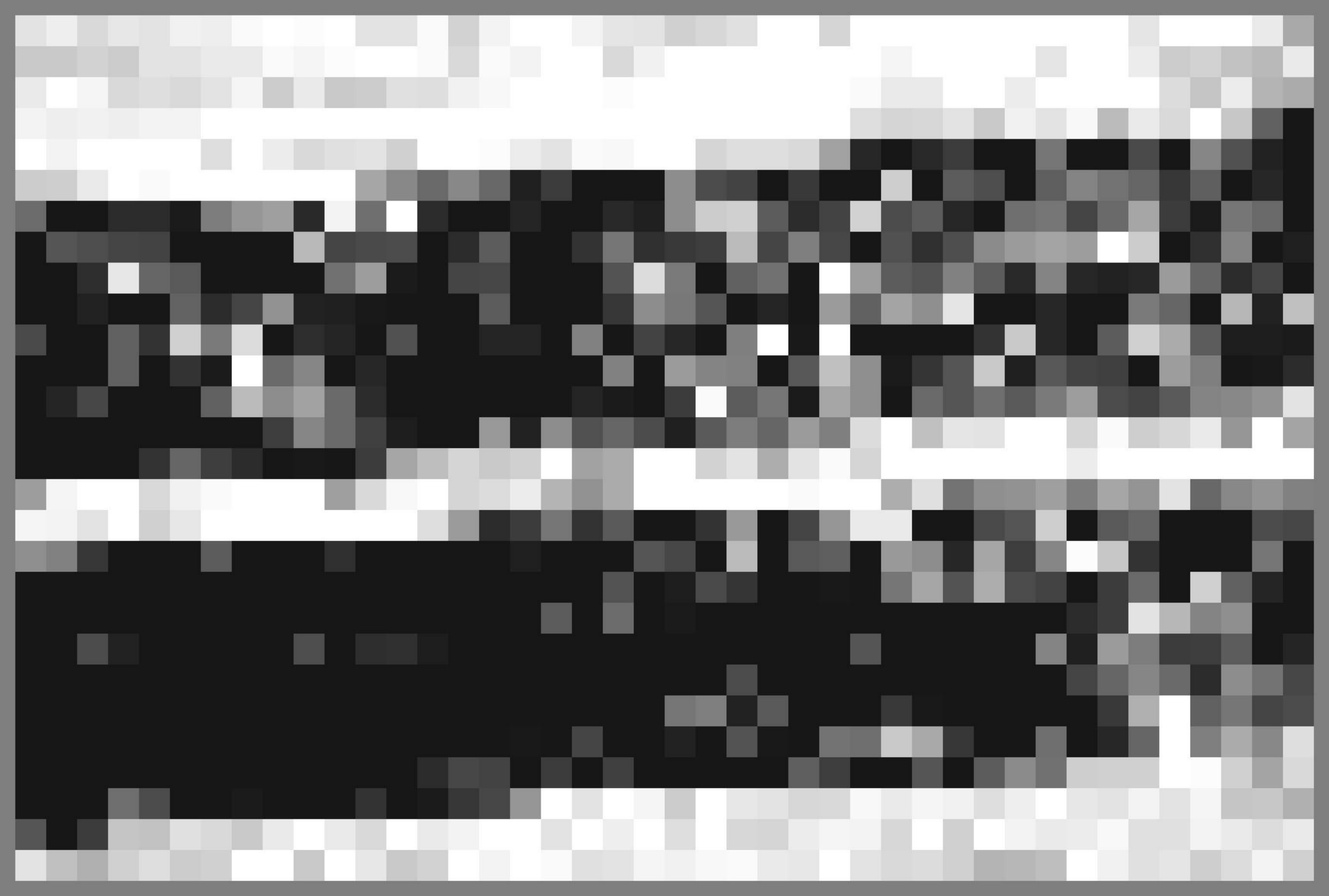}
       \subcaption{}
    \end{minipage}
    \end{minipage}
        \begin{minipage}{0.24\linewidth}
    \begin{minipage}{\linewidth}
       \centering
       \includegraphics[width=\linewidth]{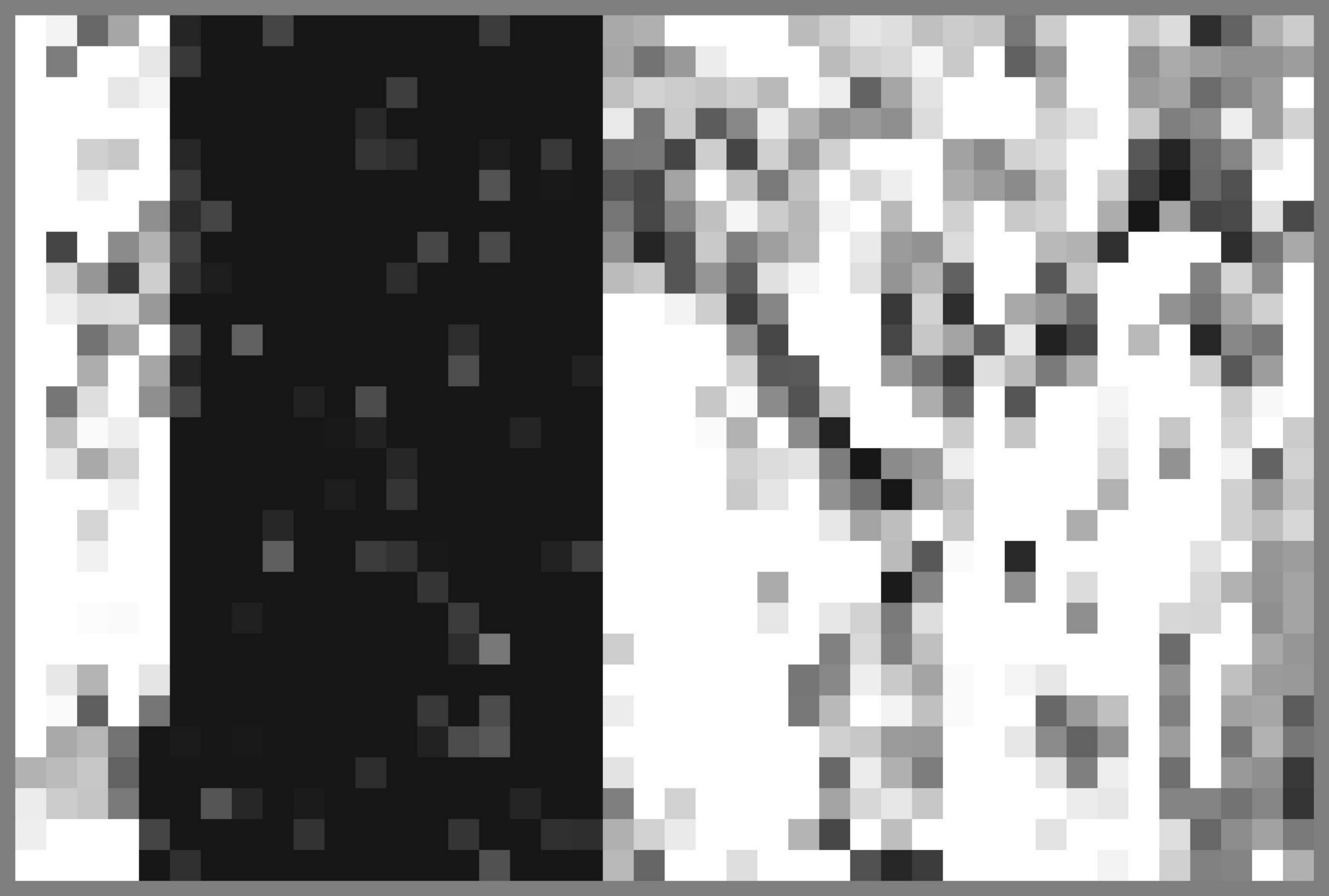}
    \end{minipage}
    \begin{minipage}{\linewidth}
       \centering
       \includegraphics[width=\linewidth]{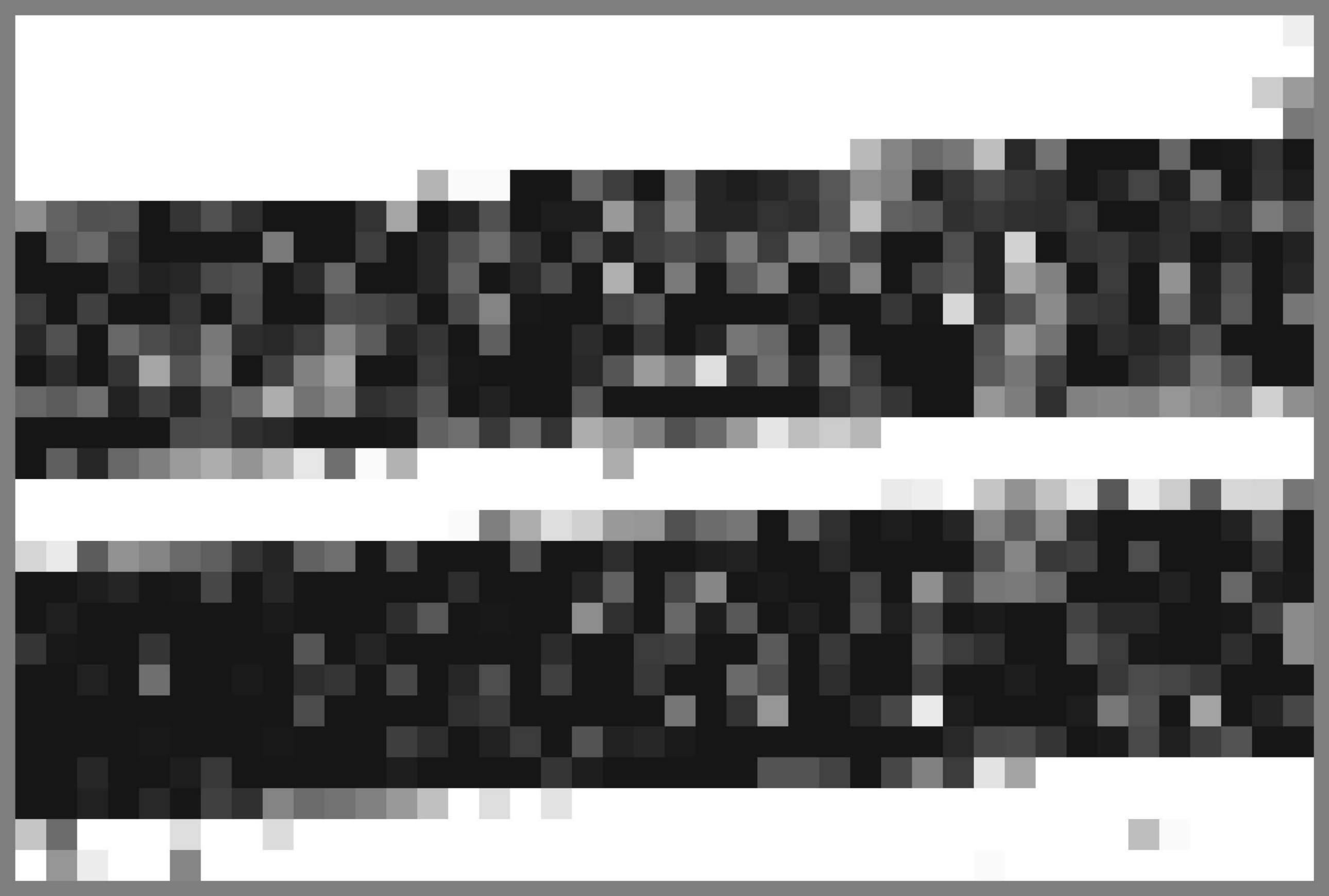}
       \subcaption{}
    \end{minipage}
    \end{minipage}
        \begin{minipage}{0.24\linewidth}
    \begin{minipage}{\linewidth}
       \centering
       \includegraphics[width=\linewidth]{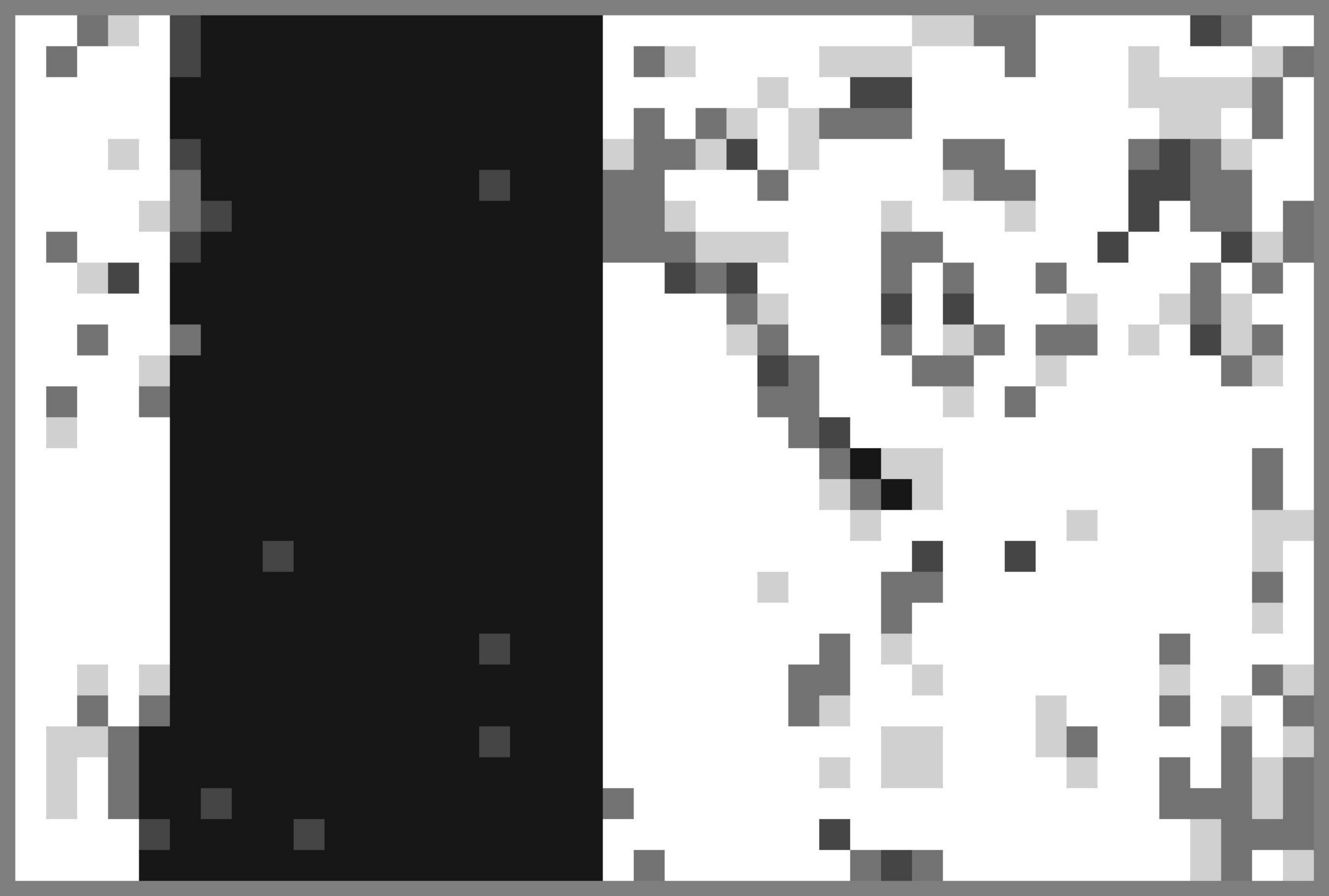}
    \end{minipage}
    \begin{minipage}{\linewidth}
       \centering
       \includegraphics[width=\linewidth]{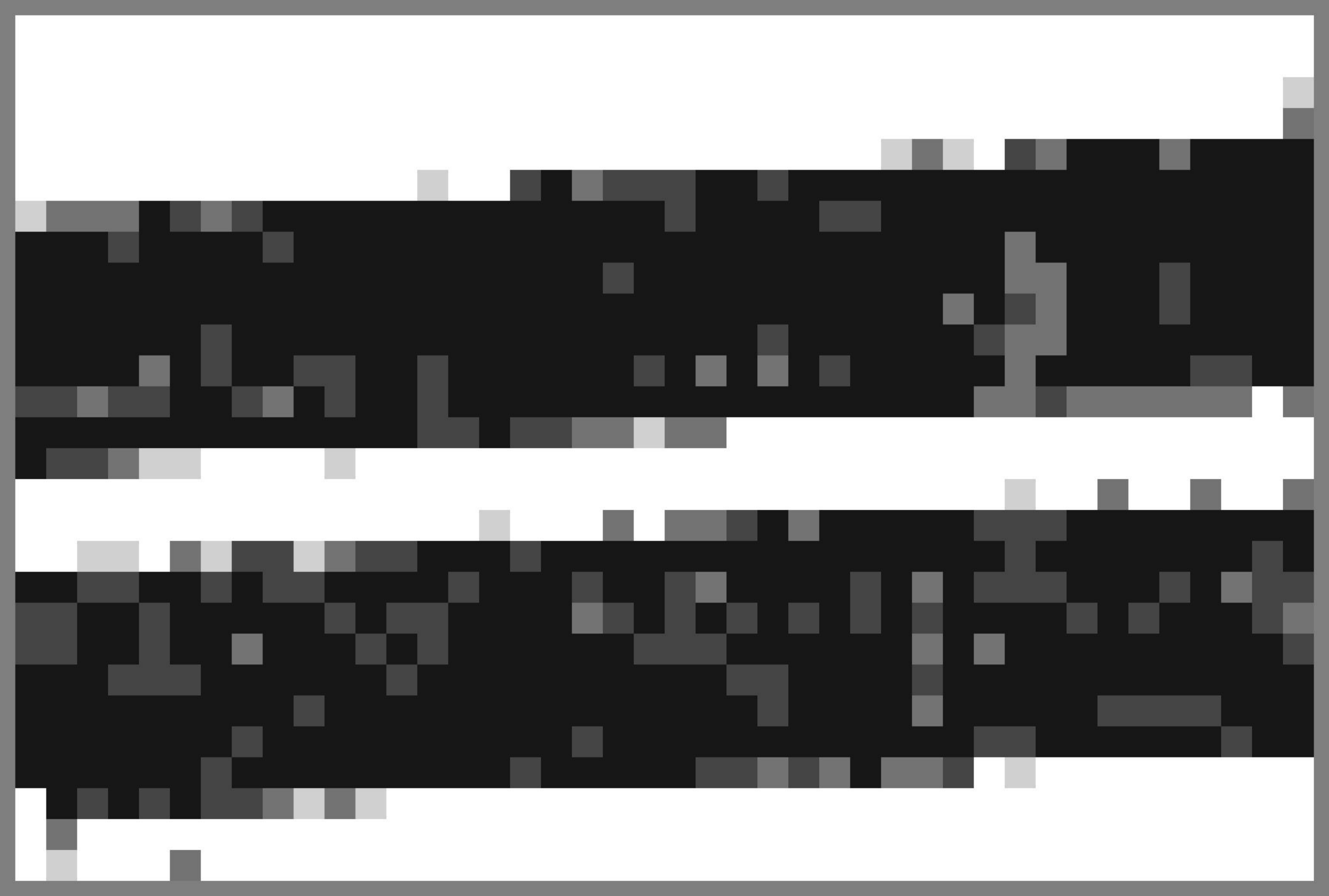}
       \subcaption{}
    \end{minipage}
    \end{minipage}
    \vspace{-1em}
    \caption{\small \it 
     Example of blur map estimation (the darker, the sharper). (a) The left view of the DP image. (b) and (c) Blur maps obtained with prompts of the blur-aware and DP-aware formats. (d) The blur maps obtained with the ensembling-based format, \ie, all eight blur-aware and DP-aware prompts are used (please refer to \cref{tab:promp} for details). 
    } 
    \vspace{-4mm}
    \label{fig:blurm}
\end{figure}

\vspace{-5mm}
\paragraph{Blur-aware Format.} An intuitive approach is setting a prompt $\mathbf{p}$ that describes the blur to query a blur map $\vM$, \eg, $\mathbf{p} =$ \texttt{[A blurry image.]}. The $\eta(\cdot,\cdot)$ is then defined to synthesize a center view of the DP pair by averaging the left view $\vB_{\rm L}$ and right view $\vB_{\rm R}$, as the average of two sub-images is equivalent to the image captured by a regular sensor \cite{punnappurath2020modeling,abuolaim2022improving}.
However, the disparity information among the DP pair is neglected, resulting in a sub-optimal blur map.  Refer to Fig.~\ref{fig:blurm} for comparisons.

\vspace{-5mm}
\paragraph{DP-aware Format.} 
{Based on the DP image formation process, the disparity is proportional to the degree of blur exhibited in the DP view \cite{abuolaim2020defocus}. We thus propose to guide CLIP to understand the disparity exhibited between $\vB_{\rm L}$ and $\vB_{\rm R}$ which eventually leads to better mask ($\vM$) generation. To this end,} we first formulate the disparity in the DP pair and then study the disparity-aware prompt and function $\eta(\cdot,\cdot)$ designing. 
Such design leads us to query degrees of displacement between the left and right view from the CLIP, without any fine-tuning, to obtain the blur map $\vM$.

\begin{figure}[!t]
    \centering
    \includegraphics{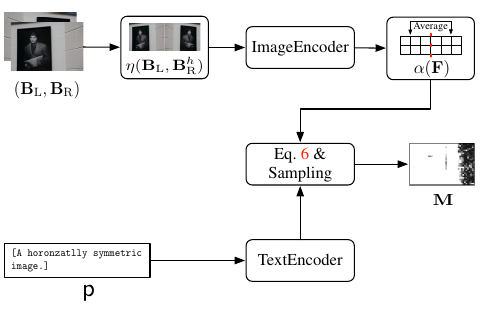}
    \vspace{-3mm}
    \caption{\small \it The pipeline of our DP-aware format for obtaining the blur map $\vM$. Given a DP pair $(\vB_{\rm L}, \vB_{\rm R})$, the function $\eta(\vB_{\rm L},\vB_{\rm R}^{h})$ concatenate the left view $\vB_{\rm L}$ and with horizontally flipped right view $\vB_{\rm R}^{h}$ for obtaining the $\vF$ from the CLIP image encoder. We then apply $\alpha(F)$ to averagely pool the feature map according to the transformation imposed on the input DP pair, resulting in a new feature embedding with the same aspect ratio of the input DP pair. Meanwhile, we set the prompts $\vp$ to describe the symmetry of $\eta(\vB_{\rm L},\vB_{\rm R}^{h})$, obtaining text embedding $\vt$. We finally have the blur map $\vM$ by computing Eq.~\ref{eq:dpmas} and performing sampling. 
    }
    \label{fig:clip}
    \vspace{-4mm}
\end{figure}

Note that the CLIP is the vision-text alignment model trained with monocular image input. To extract the displacement between left/right views of the DP pair, we explore several input-prompt formats:
i) Concatenating the left and right views horizontally, and prompting their difference,~\eg,~\texttt{[The left and right of the image are  different.]}; 
ii) Prompting the horizontal symmetry between the left and right views,~\eg,~\texttt{[A horizontally symmetric image.]} (Fig.~\ref{fig:clip}).  

Empirically, we achieve the best performance with the second prompt format. When describing the symmetry in the prompt $\vp$, we concatenate the left view $\vB_{\rm L}$ with a horizontally flipped right view $\vB_{\rm R}^h=\text{Flip}(\vB_{\rm R})$ along the width axial in the $\eta(\cdot,\cdot)$. We then obtained image features $\vF = \text{ImageEncoder}\bigl(\eta(\vB_{\rm L}, \vB_{\rm R}^h)\bigr)$. To transform  $\vF$ to the same aspect ratio of $\vB_{\rm L}$ and $\vB_{\rm R}$, 
{we average the corresponding pixel-wise features based on the transformation in $\eta(\vB_{\rm L},\vB_{\rm R}^h)$ and have $\alpha(\vF)$.}
For $ i \in \{1,\cdots, \vh_{\vw}\}$ and $j \in \{1, \cdots, \lfloor \frac{1}{2} \vw_{\mathsf{s}} \rfloor\}$, we define 
{\small
\begin{align}
    \alpha(\vF)(i,j) &= \frac{1}{2}\vF(i,j) +  \frac{1}{2} \vF(i, \vw_{\mathsf{s}}-j) \ . 
\end{align}
}

\noindent{Back} to the left-hand side of Eq.~\ref{eq:maskbar}, we have 
{\small{
\begin{align}
    \vM_{\vF}(i,j) = 1 - \vM^s_{\vF}(i,j) = 1 - \sigma\big(\texttt{sim}(\alpha(\vF)(i,j),\vt^s)\big) \ . \label{eq:dpmas}
\end{align}
}}

\noindent{Here}, $\vt^s$ is the text embedding of the prompt describing symmetry. 
{Note that the high similarity value, i.e. $\vM^s_{\vF}(i,j)=\sigma\bigl(\texttt{sim}(\alpha(\vF)(i,j),\vt^s)\bigr)$ indicates pixels in sharp regions. Therefore the blur map is derived as $1-\vM^s_{\vF}(i,j)$ in Eq.~\ref{eq:dpmas}.}
Similarly, the $\vM_{\vF}$ is sampled to the same resolution of the DP pair, returning $\vM$. 
For this blur map estimation strategy, we conclude the lemma below, 
 
\vspace{-.7em}
\begin{lemma}
A pixel is sharp if and only if it is horizontally symmetric in the left view and the horizontally flipped right view,  otherwise, the pixel is blurred.
\label{lemma:dp}
\end{lemma}
\vspace{-1em}
\begin{proof}
Following \cref{eq:disparity}, we have $\vB_{\rm L}(i,j)$ being sharp, given $\vB_{\rm L}(i,j) = \text{Flip}(\vB_{\rm R})(i, \vw-j)$. In other words, we have $\eta(\vB_{\rm L}, \vB_{\rm R}^h)(i,j) = \eta(\vB_{\rm L}, \vB_{\rm R}^h)(i, \vw_{\mathsf{s}} - j)$, demonstrating the horizontal symmetry. Reversely, the horizontal symmetry breaks once the disparity/blur occurs.  
\end{proof}
\vspace{-.5em}

\vspace{-5mm}
\paragraph{Ensembling format.}
Given the analysis that the degradations in the DP pair can be described by both the blurriness and symmetricity, we thus propose to obtain the mask by ensembling the blur-aware and DP-aware formats jointly. 
Please refer to \cref{tab:promp} for details of the 8 prompts designed from blur-aware and DP-aware formats.

\subsection{Network}
\label{sec:net}

We train an encoder-decoder based deblurring network for restoring the DP pair $(\vB_{\rm L}, \vB_{\rm R})$ into a {sharp image} $\hat{\vI}$ 
{with the derived blur map $\vM$.}
Our network consists of a backbone network based on \cite{chen2022simple}
and a BPA block to interleave with the backbone network layers. {In particular, the} BPA block uses the obtained blur map $\vM$ as deblur kernel prior to help {the} restoration of $(\vB_{\rm L}, \vB_{\rm R})$ on the fly. 

\vspace{-5mm}
\paragraph{{Self-attention.}}
The self-attention (SA) mechanism is widely used in the defocus deblurring tasks~\cite{zamir2022restormer}. We first briefly review it. Let $\vX \in \vR^{\vh_{\vx} \times \vw_{\vx} \times \vc_{\vx}}$ be the intermediate feature embedding of a blur image 
from the backbone encoder, where $\vh_{\vx}$, $\vw_{\vx}$, $\vc_{\vx}$ are height, width, and channel of the feature. The self-attention is input with query $\vQ$, key $\vK$, and value $\vV$ that are linearly projected from $\vX$, and then calculate an attention map between $\vQ$ and $\vK$ for adaptively aggregating $\vV$. This process is defined as follows,
\begin{align}
    &\vQ = \vX \vW_{\vQ} \ , \quad \vK = \vX \vW_{\vK} \ , \quad \vV = \vX \vW_{\vV} \ , \\
    &\text{Attention}(\vQ, \vK, \vV) = \text{Softmax}(\vQ \vK^{\top} / \vbeta) \vV \label{eq:att} \ ,
\end{align}
where $\vW_{\vQ}$,  $\vW_{\vK}$, $\vW_{\vV}$ $\in  \vR^{\vh_{\vx} \times \vw_{\vx} \times \vc_{\vx}}$ are learnable weight matrices, $\beta$ is a normalization factor that is typically set to the square root of the latent dimension, and $\text{Softmax}(\vQ \vK^{\top} / \vbeta) \in  \vR^{(\vh_{\vx} \times \vw_{\vx}) \times (\vh_{\vx} \times \vw_{\vx})} $ is known as the attention map. 

\begin{figure}[!t]
    \centering
    \begin{minipage}{0.24\linewidth}
       \centering
       \includegraphics[width=\linewidth,trim={120pt 60pt 0 0},clip]{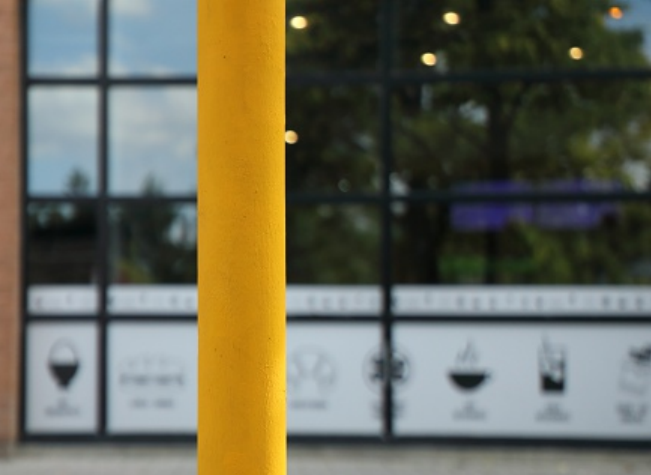}
       \subcaption{}
    \end{minipage}
    \begin{minipage}{0.24\linewidth}
       \centering
       \includegraphics[width=\linewidth,trim={120pt 60pt 0 0},clip]{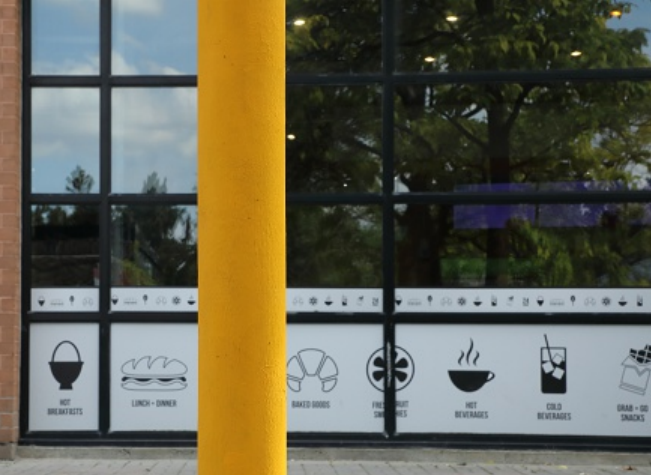}
       \subcaption{}
    \end{minipage}
    \begin{minipage}{0.24\linewidth}
       \centering
       \includegraphics[width=\linewidth,trim={120pt 60pt 0 0},clip]{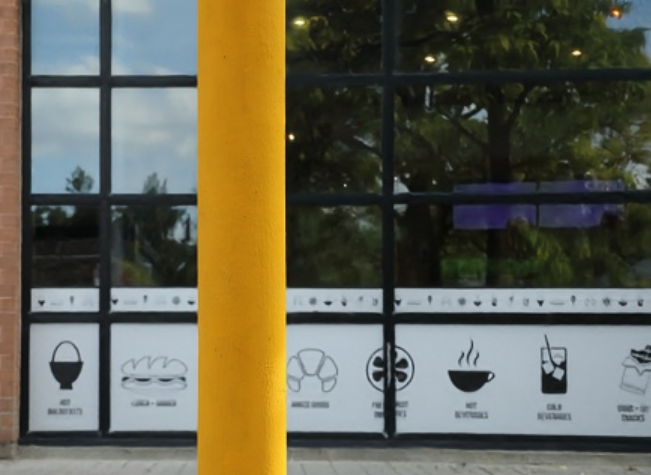}
       \subcaption{}
    \end{minipage}
    \begin{minipage}{0.24\linewidth}
       \centering
       \includegraphics[width=\linewidth,trim={120pt 60pt 0 0},clip]{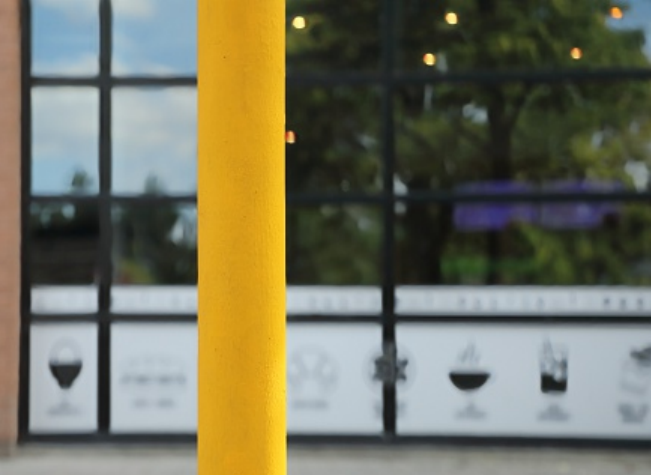}
       \subcaption{}
    \end{minipage}
    \vspace{-1em}
    \caption{\small \it Example of deblurring with the blur map projected deblur kernel $\vO$. (a) and (b) are the left view of the DP pair and the ground truth sharp image. (c) and (d) are deblurred images with and without using $\vO$. Zoom in for better quality.
    }
    \label{fig:attev}
    \vspace{-2mm}
\end{figure}

\vspace{-5mm}
\paragraph{BPA Block.} 
The attention map can be considered as an adaptive deblur kernel prior for restoring the DP pair in the feature space. We are motivated to modulate the attention map by supplying the blur map $\vM$, utilizing blur-related prior knowledge to additionally assign a $\vq \times \vq$ deblur kernel for each region in $\vX$ {to aggregate a sharp feature embedding from its neighboring regions.}
We re-write the Eq.~\ref{eq:att} as 
\begin{align}
    &\text{Attention}(\vQ, \vK, \vV) = \text{Softmax}(\vQ \vK^{\top} / \beta + \vO) \vV  \ , \\
    &\vO = \text{Pad}(\text{FFN}_{\text{Conv}}(\sigma^{-1}(\vM))) \ ,
\end{align}
and show that the restoration result is further improved in Fig.~\ref{fig:attev}, where $\text{FFN}_{\text{Conv}}(\cdot)$ is a convolution-based feed forward layer that is used to learn a $\vq \times \vq$ deblur kernel for each region of $\vX$ from the unnormalized blur map $\sigma^{-1}(\vM)$, resulting in $\text{FFN}_{\text{Conv}}(\sigma^{-1}(\vM)) \in \vR^{\vh_{\vx} \times \vw_{\vx} \times \vq^{2}}$. The $\text{Pad}(\cdot)$ function reshapes $\text{FFN}_{\text{Conv}}(\sigma^{-1}(\vM))$ into $\vR^{\vh_{\vx} \times \vw_{\vx} \times \vq \times \vq}$,  and pads the reshaped $\text{FFN}_{\text{Conv}}(\sigma^{-1}(\vM)) $ to $\vR^{(\vh_{\vx} \times \vw_{\vx}) \times (\vh_{\vx} \times \vw_{\vx})}$, for shape alignment issue.   

\begin{table*}[!t]
    \centering
    \caption{\small \it Quantitative comparisons on the DPD-blur dataset \cite{abuolaim2020defocus} composed of 37 indoor and 39 outdoor scenes. We respectively color the best and the second-best performance in \textcolor{red}{red} and \textcolor{blue}{blue}.  
    }
    \vspace{-1em}
    \label{tab:dpd}
    \setlength{\tabcolsep}{4.4pt}
    \small
    \begin{tabular}{l|cccc|cccc|cccc}
        \toprule
       \multirow{2}{*}{Method} & \multicolumn{4}{c}{\bf Outdoor Scenes} & \multicolumn{4}{c}{\bf Indoor Scenes} & \multicolumn{4}{c}{\bf Average} \\
       \cmidrule{2-13}
        & PSNR$_\uparrow$ & SSIM$_\uparrow$ &MAE$_\downarrow$ & MSE\_rel$_\downarrow$ & PSNR$_\uparrow$ & SSIM$_\uparrow$ & MAE$_\downarrow$ & MSE\_rel$_\downarrow$ & PSNR$_\uparrow$ & SSIM$_\uparrow$ & MAE$_\downarrow$ & MSE\_rel$_\downarrow$   \\
      \midrule

        EBDB &25.77	&0.772	&0.040	&0.051	&21.25	&0.599	&0.058	&0.086	&23.45	&0.683	&0.049	&0.067\\
        DMENet &25.50  &0.788  &0.038  &0.053  &21.43  &0.644  &0.063  &0.084  &23.41  &0.714  &0.051 &0.067	
 \\
        DPDNet &27.48  &0.849  &0.029  &0.042  &22.90  &0.726  &0.052  &0.071  &25.13  &0.786  &0.041 &0.055	
\\
        RDPD &28.10	&0.843	&0.027	&0.039	&22.82	&0.704	&0.053	&0.072	&25.39	&0.772	&0.040 &0.053
\\
        DDDNet &27.57 	&0.833	&0.030	&0.041	&23.28	&0.708	&0.050	&0.068	&25.36	&0.768	&0.041 &0.054
\\
        IFAN &28.66	&0.868	&0.025	&0.037	&23.46	&0.743	&0.049	&0.067	&25.99	&0.804	&0.037 &0.050
\\
        BAMBNet &28.60	&0.872	&0.026	&0.037	&24.30	&0.772	&0.045	&0.060	&26.40	&0.821	&0.036 &0.047
\\
        DeepRFT &28.48  &0.870  &0.025  &0.038   &23.09 &0.736 &0.051  &0.070  & 25.71 & 0.801 & 0.037 & 0.051\\
        Restormer  &29.48	&\textcolor{blue}{0.895}	&0.023	&0.034	&23.97	&0.773	&0.047	&0.063	&26.66	&0.833	&0.035 &0.046
\\
        K3DN  &\textcolor{blue}{29.87}	&0.890	&0.022	&\textcolor{blue}{0.032}	&24.37	&\textcolor{blue}{0.780}	&0.046	&\textcolor{blue}{0.060}	&\textcolor{blue}{27.06}	&\textcolor{blue}{0.835}	&0.034 &\textcolor{blue}{0.044}
\\
        \midrule
        \rowcolor{gray!30}
        Ours (Small) &\textcolor{black}{29.55}	&0.893	&\textcolor{blue}{0.021}	&0.033	&\textcolor{blue}{24.38}	&0.773	&\textcolor{blue}{0.043}	&0.065	&26.91	&0.831	&\textcolor{blue}{0.032}	&0.045
\\      \rowcolor{gray!30}
        Ours (Large) &\textcolor{red}{29.88}	&\textcolor{red}{0.896}	&\textcolor{red}{0.020}	&\textcolor{red}{0.032}	&\textcolor{red}{24.73}	&\textcolor{red}{0.786}	&\textcolor{red}{0.037}	&\textcolor{red}{0.059}	&\textcolor{red}{27.24}	&\textcolor{red}{0.840}	&\textcolor{red}{0.028}	&\textcolor{red}{0.043}
\\
        \bottomrule
    \end{tabular}
    \vspace{-1em}
\end{table*}

\begin{table*}[!t]
\begin{minipage}{.5\linewidth}
\caption{\small \it Quantitative comparisons on DDD-syn dataset.}
\vspace{-2em}
    \label{tab:ddd}
\setlength{\tabcolsep}{7.5pt}
\flushleft
\small
\begin{tabular}{l|cccc}
    \toprule
         Method & PSNR$_\uparrow$ & SSIM$_\uparrow$ & MAE$_\downarrow$ & MSE\_rel$_\downarrow$    \\
        \midrule
        EBDB      & 26.48    &  0.683     &  0.036  &  0.047\\ 
        DMENet    & 30.14    &  0.939     &  0.017  &  0.031 \\ 
        DPDNet    & 31.45    &  0.926     &  0.016  &  0.027\\
        RDPD      & 32.42    &  0.912     &  0.016  &  0.024\\
        DDDNet    & 33.21    &  0.956     &  0.010  &  0.022\\
        IFAN      & 34.18    &  0.929     &  0.012  &  0.020\\
        BAMBNet   & 35.90    &  0.954     &  0.012  &  0.016\\
        DeepRFT   & 36.52    & 0.952 & 0.010 & 0.015 \\
        Restormer & 36.44    &  0.957     &  0.010  &  0.015\\
        K3DN & \textcolor{blue}{38.23}    &  \textcolor{blue}{0.965}     &  \textcolor{blue}{0.008}  &  \textcolor{blue}{0.012}\\
        \midrule
        \rowcolor{gray!30}
        Ours (Small) &37.27	&0.959	&0.010	&0.014\\
        \rowcolor{gray!30}
        Ours (Large) &\textcolor{red}{38.62}	&\textcolor{red}{0.971}	&\textcolor{red}{0.008}	&\textcolor{red}{0.012}\\
        \bottomrule
\end{tabular}
\end{minipage}
\hfill
\begin{minipage}{.5\linewidth}
\caption{\small \it Quantitative comparisons on RDPD dataset.}
\vspace{-2em}
\label{tab:rdp}
\setlength{\tabcolsep}{7.5pt}
\flushright
\small
\begin{tabular}{l|cccc}
        \toprule
         Method & PSNR$_\uparrow$ &SSIM$_\uparrow$ & MAE$_\downarrow$ & MSE\_rel$_\downarrow$  \\
        \midrule
        EBDB       &  24.23      &  0.637   &  0.031  &  0.054    \\
        DMENet     &  25.17      &  0.731   &  0.036  &  0.061   \\
        DPDNet     &  29.84      &  0.828   &  0.032  &  0.055    \\
        RDPD       &  31.09      &  0.861   &  0.025  &  0.028    \\
        DDDNet     &  30.14      &  0.840   &  0.016  &  0.031   \\
        IFAN       &  31.12      &  0.865   &  0.023  &  0.028    \\
        BAMBNet    &  31.78      &  0.867   &  0.016  &  0.026   \\
        DeepRFT &  31.71 & 0.879 &0.016& 0.024\\
        Restormer  &  32.27      &  0.871   &  0.016  &  0.024    \\
        K3DN  &  32.84      &  \textcolor{blue}{0.905}   &\textcolor{blue}{0.014}  &  0.023   \\
        \midrule
        \rowcolor{gray!30}
        Ours (Small) &\textcolor{blue}{32.98}	&\textcolor{black}{0.895}	&\textcolor{blue}{0.014}	&\textcolor{blue}{0.022}\\
        \rowcolor{gray!30}
        Ours (Large) &\textcolor{red}{33.44}	&\textcolor{red}{0.907}	&\textcolor{red}{0.013}	&\textcolor{red}{0.021}\\
        \bottomrule
\end{tabular}
\end{minipage}
\vspace{-1em}
\end{table*}

\subsection{Network Training}
\label{sec:los}
We optimize our deblurring network with the widely used Charbonnier loss. Moreover, we propose a blur-aware loss and a blur weighting loss to penalize the blurred regions in the restoration $\hat{\vI}$. 

\vspace{-5mm}
\paragraph{Charbonnier loss.} We encourage the restoration to be close to the ground truth by using Charbonnier loss. We empirically set $\varepsilon$ to $10^{-4}$, and define
\begin{equation}
{{\vL}_{\text{char}}}= \sqrt{\lVert \vI(i,j) - \hat{\vI}(i,j) \rVert^{2}+ \varepsilon^{2}} \ . \label{eq:charb}
\end{equation}

\vspace{-5mm}
\paragraph{Blur Weighting Loss.} We impose an adaptive penalty on the blurred regions of the DP pair $(\vB_{\rm L}, \vB_{\rm R})$, considering the degree of blur measured by CLIP. It encourages  the network to focus on the restoration of heavily blurred regions. We first normalize the deviations between $\hat{\vI}$ and $\vI$ by the deviations between $(\vB_{\rm L}, \vB_{\rm R})$ and $\vI$, and then re-use the blur map to select blurred regions residing in the DP pair.
\begin{align}
    \vL_{\text{bwl}} = \frac{\lVert \vI(i,j) - \hat{\vI}(i,j) \rVert^{2}}{\lVert \hat{\vI}(i,j) - \frac{1}{2}\bigl(\vB_{\rm L}(i,j) + \vB_{\rm R}(i,j)\bigr) \rVert^{2}} \cdot  \vM(i,j) \ ,
\end{align}
where we use the blur map $\vM$ (with soft logits) to reflect the blur levels detected by CLIP.

\vspace{-5mm}
\paragraph{Blur-aware Loss.} We encourage  $\hat{\vI}$ to be sharp by leveraging the versatile knowledge contained in the CLIP. We generate the blur map $\hat{\vM}$ by the blur-aware format for the restoration, and penalize the blur exhibited in $\hat{\vI}$. Similar to $\vL_{\text{bwl}}$, we use the unnormalized version of the blur map,
\begin{align}
    \vL_{\text{bal}} =  \hat{\vM}(i,j) \ . 
\end{align}
This encourages $\hat{\vM}$ to be zero, \ie, no blur exists in the restoration.

\vspace{-5mm}
\paragraph{Total Loss.} Our network is optimized by minimizing the following loss function 
\begin{equation}
\vL_{\text{total}}= \frac{1}{\lvert \Omega \rvert }\sum_{(i,j) \in \Omega} \left(\vL_{\text{char}} + \lambda_{1}  {\vL}_{\text{bal}} + \lambda_{2}   \vL_{\text{bwl}}\right) \ ,
\end{equation}
where $\frac{1}{\lvert \Omega \rvert }$ is a normalization factor determinated by the image domain size $\lvert \Omega \rvert$, and $\lambda_{1}$ and $\lambda_{2}$ are hyperparameters to balance losses.

\section{Experiments and Analysis}

\paragraph{Implementation Details.}
We implement our model under the PyTorch framework. We use the pre-trained and ViT-B/32 \cite{dosovitskiy2020image} architecture-based CLIP \cite{radford2021learning} for our blur map estimation strategy. We train the deblurring network in two stages for $6 \times 10^{6}$ steps in total with the Adam optimizer. In each stage, the learning rate is decayed from $2 \times 10^{-4}$ to $10^{-6}$ by a cosine annealing strategy. In the first stage, we train the network for $5 \times 10^{6}$ steps, and only use the Charbonnier loss $\vL_{\text{char}}$. In the second stage, the network is finetuned for $1 \times 10^{6}$ steps with $\vL_{\text{total}}$, where we have $\lambda_{1} = 10^{-1}$ and $\lambda_{2} = 2 \times 10^{-1}$. More implementation details are given in our supplementary material.

\vspace{-5mm}
\paragraph{Datasets.} Our method is evaluated on three real-world DP defocus deblurring datasets, DPD-blur dataset \cite{abuolaim2020defocus} and DPD-disp dataset \cite{punnappurath2020modeling}, and PixelDP dataset \cite{xin2021defocus}.
We follow \cite{abuolaim2020defocus} on the training and testing splits. For the DPD-disp dataset and PixelDP dataset, we use them for testing only, re-using the checkpoints trained from the DPD-blur dataset as they do not provide ground-truth (GT) data or with limited dataset size. Moreover, two synthetic datasets, DDD-syn \cite{pan2021dual} and RDPD \cite{xin2021defocus}, are also used for evaluations.

\vspace{-5mm}
\paragraph{Evaluation Metrics.}
We use peak signal-to-noise ratio (PSNR), structural similarity (SSIM), mean absolute error (MAE), and relative mean squared error (MSE\_rel) as our evaluation metrics.

\begin{figure*}[!t]
    \centering
    \begin{tikzpicture}
    \node[anchor=south west,inner sep=0] (image) at (0,0) {\includegraphics[width=.78\linewidth]{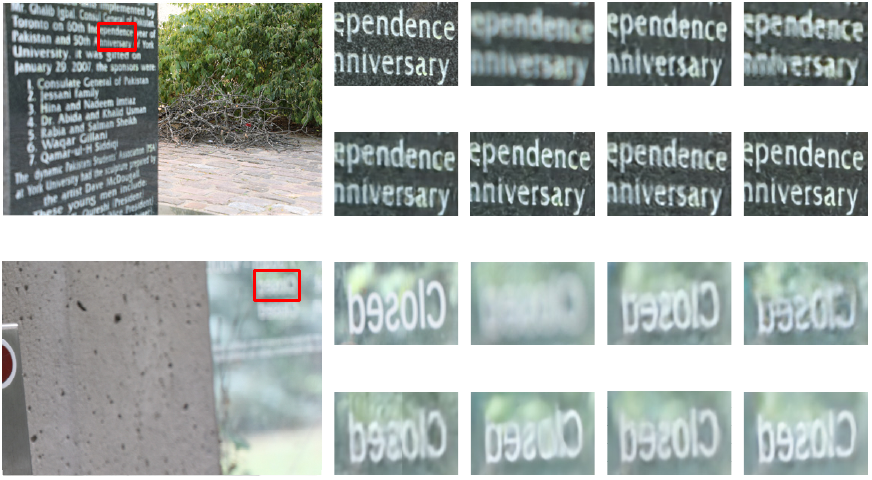}};
    \begin{scope}[x={(image.south east)},y={(image.north west)}]
        \draw (0.2,  -0.04) node {Blurry Image};
        \draw (0.2,  0.5) node {Blurry Image};
    \foreach \r in {0,...,1}
        \foreach \mylabel [count=\x from 0] in {DeepRFT \cite{deeprft}, Restormer \cite{zamir2022restormer}, K3DN \cite{yang2023k3dn}, Ours}
            \draw (0.45 + \x * 0.162, \r * 0.54 - 0.04) node {\mylabel};
    \foreach \r in {0,...,1}
        \foreach \mylabel [count=\x from 0] in {GT Image, Blurry Image, IFAN \cite{lee2021iterative}, DDDNet \cite{pan2021dual}}
            \draw (0.45 + \x * 0.157, \r * 0.54 + 0.27 - 0.04) node {\mylabel};
    \end{scope}
    \end{tikzpicture}
    \vspace{-1em}
    \caption{\small \it Qualitative comparison on the DPD-blur dataset \cite{abuolaim2020defocus}. We present the blurry image ($\vB_{\rm L}$) in the first column, and the ground truth (GT) image of regions residing in the red bounding box is shown in the second column. The cropped blurred image of the red bounding box region is in the third column.
    }
    \label{img:compa}
    \vspace{-1.0em}
\end{figure*}
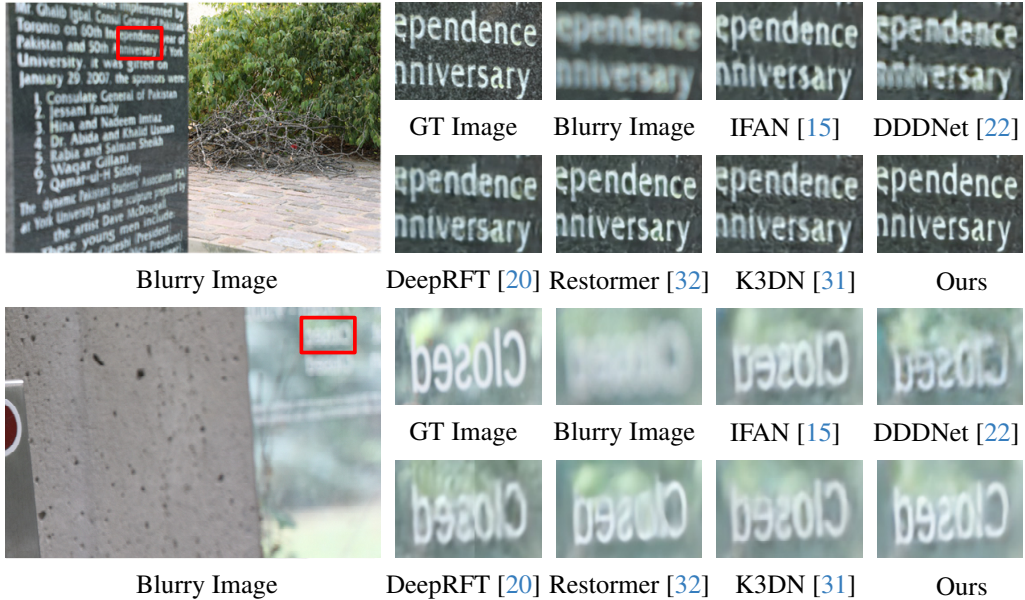

\vspace{-5mm}
\paragraph{Baseline Methods.} We compare with SOTA DP defocus deblurring methods, \ie, EBDB \cite{karaali2017edge}, DMENet \cite{lee2019deep}, DPDNet \cite{abuolaim2020defocus}, RDPD \cite{abuolaim2021learning}, DDDNet \cite{pan2021dual}, IFAN \cite{lee2021iterative}, BAMBNet \cite{liang2021bambnet}, DeepRFT \cite{deeprft}, Restormer \cite{zamir2022restormer}, and K3DN \cite{yang2023k3dn}.

\subsection{Comparisons with State-of-the-Art}
\paragraph{Quantitative Comparison.} We compare with the state-of-the-art methods on the DPD-blur, DDD-syn, and RDPD datasets in \cref{tab:dpd}, \cref{tab:ddd}, and \cref{tab:rdp}, respectively. We train our model with small and large configurations, and details are given in our supplementary material.  Our observations are as follows:
i) With our model (Small), we achieve similar restoration results with Restormer and K3DN, the third-best and second-best method, in all datasets. ii) When scaling to the large model, we consistently outperform all methods on all evaluation metrics. For example, our method, K3DN, and Restormer respectively achieve 27.24/27.06/26.66, 38.62/38.23/36.44, and 33.44/32.84/32.27 PSNR (dB) in the DPD-blur, DDD-syn, and RDPD datasets. Meanwhile, on a single RTX 4090 GPU, we remain to have a lower Infer. time (considering the frozen parameters in CLIP) which is 0.042s and 0.054s for our small and large models, while K3DN/Restormer takes 0.198s/0.255s.

\begin{figure}[!h]
    \vspace{-0.5em}
    \scriptsize
    \centering
    \begin{minipage}{0.24\linewidth}
       \centering
       \includegraphics[width=\linewidth,trim={0 0 0 0},clip]{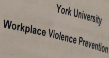}
       \vspace{-1.5em}
       \subcaption{Input}
    \end{minipage}
    \begin{minipage}{0.24\linewidth}
       \centering
       \includegraphics[width=\linewidth,trim={0 0 0 0},clip]{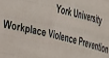}
       \vspace{-1.5em}
       \subcaption{K3DN}
    \end{minipage}
    \begin{minipage}{0.24\linewidth}
       \centering
       \includegraphics[width=\linewidth,trim={0 0 0 0},clip]{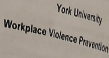}
       \vspace{-1.5em}
       \subcaption{Restormer}
    \end{minipage}
    \begin{minipage}{0.24\linewidth}
       \centering
       \includegraphics[width=\linewidth,trim={0 0 0 0},clip]{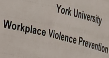}
       \vspace{-1.5em}
       \subcaption{Ours}
    \end{minipage}\\
    \vspace{-1.1em}
    \caption{\small \it An example of the restoration results on the sharp region of the input DP image (left view). 
    }
    \vspace{-0.5em}
    \label{img:sharpblur}
\end{figure}

\vspace{-5mm}
\paragraph{Qualitative Comparison.} We compare with state-of-the-art methods on the DPD-blur dataset (Fig.~\ref{img:compa}). As shown, our method performs sharper and clearer restorations of edges and texts. 
To further demonstrate the benefits of using CLIP to generate the blur mask unsupervisedly in our approach, we provide an example in Fig.~\ref{img:sharpblur}, which preserves sharp regions (\eg, text ‘York’) in the input blur image effectively compared to the second-best baseline. 

\begin{table*}[!t]
\begin{minipage}{\linewidth}
 \small 
 \setlength{\tabcolsep}{16.5pt}
 \centering
  \caption{\small \it Prompts used for blur map estimation.}
  \vspace{-1em}
    \begin{tabular}{lll}
    \toprule
     & \multicolumn{1}{c}{Blur-aware Prompt} & \multicolumn{1}{c}{DP-aware Prompt} \\
    \midrule
    \multicolumn{1}{c}{$\vp_{1}$} & \texttt{[A blurry image.]} & \texttt{[A symmetrical image.]} \\
    \multicolumn{1}{c}{$\vp_{2}$} & \texttt{[An obscured image.]}  & \texttt{[A horizontally symmetrical image.]}\\
    \multicolumn{1}{c}{$\vp_{3}$} & \texttt{[An out of focus image.]} & \texttt{[A left-right symmetrical image.]}   \\
    \multicolumn{1}{c}{$\vp_{4}$} & \texttt{[An image lacking sharpness.]} & \texttt{[A symmetrical image at the pixel level.]}\\
    \bottomrule
    \end{tabular}
  \label{tab:promp}
  \vspace{.5em}
\end{minipage}
\begin{minipage}{\linewidth}
      \centering
     \caption{\small \it Restoration performance of our methods by using blur-aware or/and DP-aware prompts for blur map estimation. Ens.$^{1}$ denotes ensembling the group of blue-aware prompts and DP-aware prompts individually, and Ens.$^{2}$ ensembles all prompts.
     }
     \vspace{-1em}
     \small 
     \setlength{\tabcolsep}{13.3pt}
    \begin{tabular}{l|cccc|cccc}
    \toprule
     & \multicolumn{4}{c|}{Blur-aware Prompt} & \multicolumn{4}{c}{DP-aware Prompt} \\
     \cmidrule{2-9}
     & PSNR$_\uparrow$ & SSIM$_\uparrow$ & MAE$_\downarrow$ & MSE\_rel$_\downarrow$ & PSNR$_\uparrow$ & SSIM$_\uparrow$ & MAE$_\downarrow$ & MSE\_rel$_\downarrow$ \\
     \midrule
     \multicolumn{1}{c}{$\vp_{1}$} &26.83	&0.829	&0.033	&0.045	&26.82	&0.830	&0.033	&0.046\\
     \multicolumn{1}{c}{$\vp_{2}$} &26.81	&0.829	&0.033	&0.046	&26.83	&0.830	&0.032	&0.046\\
     \multicolumn{1}{c}{$\vp_{3}$} &26.75	&0.828	&0.034	&0.046	&26.82	&0.830	&0.033	&0.046\\
     \multicolumn{1}{c}{$\vp_{4}$} &26.77	&0.829	&0.033	&0.046	&26.85	&0.830	&0.032	&0.045\\
     \multicolumn{1}{c}{Ens.$^{1}$} &26.80	&0.829	&0.033	&0.046 &26.84	&0.830	&0.032	&0.046\\
     \multicolumn{1}{c}{Ens.$^{2}$} & \textbf{26.91}	&\textbf{0.831}	&\textbf{0.032}	&\textbf{0.045} &\textbf{26.91}	&\textbf{0.831}	&\textbf{0.032}	&\textbf{0.045}\\
    \bottomrule
    \end{tabular}
    \label{tab:peval}
\end{minipage}
\vspace{-.5em}
\end{table*}

\begin{table}[!t]
  \centering
  \caption{\small \it Ablation of model components.}
  \vspace{-1em}
   \small 
   \setlength{\tabcolsep}{7.3pt}
    \begin{tabular}{lcccc}
    \toprule
     Setting &  PSNR$_\uparrow$ & SSIM$_\uparrow$ & MAE$_\downarrow$ & MSE\_rel$_\downarrow$   \\
    \midrule
    i) NAFNet & 26.21	&0.817	&0.038	&0.049\\
    ii) Concat  &26.76	&0.826	&0.033	&0.046  \\
    iii) SA & 26.59 & 0.824 & 0.036 & 0.047 \\
    iv) Concat + SA & 26.80 & 0.829 & 0.032 & 0.046 \\
    v) BPA block &\textbf{26.91}	&\textbf{0.831}	&\textbf{0.032}	&\textbf{0.045}\\
    \bottomrule
    \end{tabular}
  \label{tab:archi}
\end{table}

\begin{table}[!t]
  \centering
  \caption{\small \it The effectiveness  of losses.}
  \vspace{-1em}
  \small
  \setlength{\tabcolsep}{3.5pt}
    \begin{tabular}{clcccc}
    \toprule
     BPA block & Loss &  PSNR$_\uparrow$ & SSIM$_\uparrow$ & MAE$_\downarrow$ & MSE\_rel$_\downarrow$   \\
    \midrule
    \xmark & ${\vL}_{\text{char}}$ &26.21	&0.817	&0.038	&0.049\\
    \xmark & $\vL_{\text{total}}$&26.53	&0.822	&0.036	&0.047  \\
    \midrule
    \cmark & ${\vL}_{\text{char}}$   &26.71	&0.825	&0.034	&0.046\\
    \cmark & ${\vL}_{\text{char}}$ + ${\vL}_{\text{bwl}}$  &26.76	&0.828	&0.033	&0.046\\
    \cmark & ${\vL}_{\text{char}}$ + ${\vL}_{\text{bal}}$ &26.81	&0.829	&0.032	&0.046\\
    \cmark & ${\vL}_{\text{total}}$
    &\textbf{26.91}	&\textbf{0.831}	&\textbf{0.032}	&\textbf{0.045}\\
    \bottomrule
    \end{tabular}
  \vspace{-4mm}
  \label{tab:losse}
\end{table}

\vspace{-3mm}
\subsection{Ablation Study}
We study model architecture, blur map estimation, and losses in this section. All experiments are performed in the DPD-blur dataset.

\vspace{-5mm}
\paragraph{Ablation of Model Architecture.} We study the effective usage of blur map as prior knowledge for DP defocus deblurring (Tab.~\ref{tab:archi}). To show the benefits of using blur maps, we consider four baselines:
i) NAFNet \cite{chen2022simple}, the baseline model;
ii) Concat, concatenating the blur map with the DP pair before feeding to the model;
iii) SA, using the self-attention layer \cite{dosovitskiy2020image} in Eq.~\ref{eq:att} to the model only;
iv) Concat+SA, combines the Concat and SA settings;
v) BPA block, using kernel priors from the blur map to assist deblurring. Our model consistently outperforms all baselines, which indicates the effectiveness of using the blur map. 

\vspace{-5mm}
\paragraph{Ablation of Losses.} We investigate the proposed blur weighting loss $\vL_{\text{bwl}}$ and blur-aware loss $\vL_{\text{bal}}$ (Tab.~\ref{tab:losse}) with a baseline as the $\vL_{\text{char}}$ setting. We first use $\vL_{\text{bwl}}$ and $\vL_{\text{bal}}$ to our baseline, NAFNet (\ie, not using our BPA block), that only uses $\vL_{\text{char}}$, having 0.32 PSNR (dB) improvements. Secondly, we train our model (backbone + BPA block) with Charbonnier loss, resulting in 26.71/0.825 higher PSNR (dB)/SSIM. By using the $\vL_{\text{bwl}}$ or $\vL_{\text{bal}}$, we respectively achieve 0.05/0.003 or 0.10/0.004 higher PSNR (dB)/SSIM. Combining all losses, we have the best restoration performance at 26.91/0.831 for PSNR (dB)/SSIM. Moreover, we verify the generalization ability of our $\vL_{\text{bwl}}$ and $\vL_{\text{bal}}$ losses on state-of-the-art networks, improving PSNR of Restormer and DeepRFT by 0.25 and 0.41 PSNR (dB). Refer to our supplementary materials for details.

\vspace{-4mm}
\paragraph{Ablation of Blur Map Estimation.} We ablate with different prompts used to generate the blur map. 
The blur-aware and DP-aware prompts are presented in Tab.~\ref{tab:promp}, and we show their impact on model performance with Tab.~\ref{tab:peval}.~We then ensemble the groups of blur-aware prompt and disparity-aware prompt individually (Ens.$^{1}$), leading to 26.80dB/0.829 and 26.84/0.830 PSNR (dB)/SSIM. By ensembling all prompts (Ens.$^{2}$), we have 26.91/0.831 PSNR (dB)/SSIM, consistently outperforming other approaches of generating blur masks. More ablations and analyses of CLIP can be found in our supplementary material.

\subsection{Discussion and Limitation}
\label{sec:discu}
\paragraph{Discussion.} The reader may wonder about the performance of our unsupervised blur map estimation strategy. Therefore, we perform two qualitative evaluations on the DPD-disp dataset, where the GT disparity is included. We compare with BAMBNet \cite{liang2021bambnet} and DPD-disp \cite{punnappurath2020modeling}. 

First, we consider blur map estimation. To obtain the GT blur map, the pixel with disparity $d$ that satisfies $\lvert d \rvert \leq 1$ is used as the sharp pixel, reversely, it is a burred pixel. Our method, BAMBNet and DPD-disp \cite{liang2021bambnet} respectively have 89.8\%/70.2\%/88.3\% accuracy (the higher, the better, details are in the supplementary material), showing the effectiveness of our method.

Second, we explore disparity estimation. The logits of our unnormalized blurred map are potentially linearly correlated with the ground-truth disparity. Therefore, we compare with the state-of-the-art DP disparity estimation method, DPD-disp \cite{punnappurath2020modeling}. Following \cite{punnappurath2020modeling}, we use AI(1), AI(2), and $1 - \lvert\rho_{s}\rvert$ as the evaluation metrics (the lower, the better). The AI(1), AI(2), and $1 - \lvert\rho_{s}\rvert$ of our method, BAMBNet, and DPD-disp are 0.063/0.106/0.069, 0.130/0.184/0.096, 0.306/0.772/0.0159. Though our blur map is estimated with a single forward pass, we still achieve a reasonable performance in the disparity estimation. This potentially explains the high performance of our DP defocus deblurring network, as obtaining the disparity prior benefits the defocus task.

\vspace{-5mm}
\paragraph{Limitation.} As the image encoder of CLIP is trained under a classification-alike pre-training task, it will focus more on local details that help identify its scene class. Textureless regions would produce non-significant semantic  reactions, \eg, large areas of wall, ceiling, or floor, and reduce the quality of the estimated blur map. It is possible to use a vision language framework that drives more sensitivity to regions with less texture.

\vspace{-5mm}
\paragraph{More.} In the supplementary material, more qualitative and quantitative evaluations, ablation studies, and analysis are provided, as well as results that our framework is being generalized to restoring single images with defocus blur or motion blur tasks.

\section{Conclusion and Broader Impact}
In this paper, we study end-to-end defocus deblurring of DP images, leveraging the blur-related prior knowledge from the CLIP. We first estimate a blur map with an ensemble of blur-aware and DP-aware strategies, and then use the estimated blur map as the deblurring kernel prior to restoring the DP image. We also propose a blur-aware and blur-weighting loss to regularize the restorations of the DP images during training, by distilling the knowledge of blur from CLIP.  In extensive experiments, our method outperforms past works by a large margin in quantitative and qualitative restoration performance.

\vspace{-5mm}
\paragraph{Broader Impact.} The proposed DP-aware strategy is promising to apply and extend CLIP to diverse zero-shot stereo vision tasks. We hope it will motivate future work. 


{
    \small
    \bibliographystyle{ieeenat_fullname}
    \bibliography{main}
}

\end{document}